\documentclass{article}

\usepackage{arxiv}

\usepackage[utf8]{inputenc} 
\usepackage[T1]{fontenc}    
\usepackage{hyperref}       
\usepackage{url}            
\usepackage{booktabs}       
\usepackage{nicefrac}       
\usepackage{microtype}      
\usepackage{lipsum}		
\usepackage{graphicx}
\usepackage{natbib}
\usepackage{doi}

\usepackage{amsmath,amssymb,amsfonts}
\usepackage{algorithmic}
\usepackage{graphicx}
\usepackage{textcomp}
\usepackage{amsthm}
\usepackage{url}

\usepackage[group-separator={,},group-minimum-digits=4]{siunitx}

\usepackage{amsmath,amssymb,amsfonts}
\usepackage{algorithmic}
\usepackage{graphicx}
\usepackage{textcomp}

\usepackage[utf8]{inputenc} 
\usepackage[T1]{fontenc}    
\usepackage{hyperref}       
\usepackage{url}            
\usepackage{booktabs}       
\usepackage{nicefrac}       
\usepackage{microtype}      
\usepackage{xcolor}         

\usepackage{multirow}

\usepackage{threeparttable}

\usepackage{float}
\usepackage{placeins}
\usepackage{lipsum}		
\usepackage{doi}

\title{A baseline for machine-learning-based hepatocellular carcinoma diagnosis using multi-modal clinical data}

\author{Binwu Wang
	\And
    Isaac Rodriguez
    \And
    Leon Breitinger 
    \And
    Fabian Tollens 
    \And
    Timo Itzel 
    \And
    Dennis Grimm 
    \And
    Andrei Sirazitdinov 
    \And
    Matthias Frölich 
    \And 
    Stefan Schönberg 
    \And
    Andreas Teufel
	\And  	
	Jürgen Hesser	
\And
Wenzhao Zhao	
\thanks{Binwu Wang is with Mannheim Institute for Intelligent Systems in Medicine, Medical Faculty Mannheim, Heidelberg University.
    Isaac Rodriguez, Leon Breitinger, Fabian Tollens, and Timo Itzel are with UMM Mannheim, Mannheim, Germany.
    Dennis Grimm is with Complex data processing in medical informatics (CMI), Mannheim Medical Faculty, Heidelberg University.
    Andrei Sirazitdinov is with Mannheim Institute for Intelligent Systems in Medicine, Medical Faculty Mannheim, Heidelberg University.
    Matthias Frölich and Stefan Schönberg are with Clinic for Radiology and Nuclear Medicine, Medical Faculty Mannheim,
Heidelberg University, Mannheim, Germany.
    Andreas Teufel is with Heidelberg University, Mannheim, Germany.
		Jürgen Hesser is with Interdisciplinary Center for Scientific Computing,
		Central Institute for Computer Engineering,
		CSZ Heidelberg Center for Model-Based AI, Data Analysis and Modeling in Medicine, Mannheim Institute for Intelligent Systems in Medicine, Medical Faculty Mannheim, Heidelberg University.
        Wenzhao Zhao is with School of Information Engineering, Nanjing University of Finance and Economics and Interdisciplinary Center for Scientific Computing, Mannheim Institute for Intelligent Systems in Medicine, Medical Faculty Mannheim, Heidelberg University. E-mail: zhaowenzhaoyz@163.com.
	}
}

\renewcommand{\shorttitle}{QHCC}

\begin{document}
\maketitle
\begin{abstract}
The objective of this paper is to provide a baseline for performing multi-modal data classification on a novel open multimodal dataset of hepatocellular carcinoma (HCC), which includes both image data (contrast-enhanced CT and MRI images) and tabular data (the clinical laboratory test data as well as case report forms). TNM staging is the classification task.
Features from the vectorized preprocessed tabular data and radiomics features from contrast-enhanced CT and MRI images are collected. Feature selection is performed based on mutual information. An XGBoost classifier predicts the TNM staging and it shows a prediction accuracy of $0.89 \pm 0.05$ and an AUC of $0.93 \pm 0.03$. 
The classifier shows that this high level of prediction accuracy can only be obtained by combining image and clinical laboratory data and therefore is a good example case where multi-model classification is mandatory to achieve accurate results.

\end{abstract}
\keywords{ Hepatocellular carcinoma \and multi-modal clinical data \and radiomics feature.}

\section{Introduction}
\label{sec:introduction}
Hepatocellular carcinoma (HCC) is one of the most common primary liver cancers and is associated with high cancer mortality\cite{balogh2016hepatocellular}. The clinical diagnosis of HCC typically involves data from a variety of sources. Combining information from multiple sources is helpful for a more robust and accurate diagnosis.

Image data is an important source of information for HCC diagnosis\cite{hennedige2013imaging}. Dynamic multiphase contrast-enhanced CT and MRI are considered the current standard for imaging diagnosis of HCC.
In addition to image data, tabular data such as clinical laboratory test data and case report forms are another important source of information. These data can be extracted from electronic health record systems. These data, collected over a long period of time, allow the physician to make a comprehensive assessment of the patient's status\cite{abhyankar2012standardizing,patridge2018research}.

In recent years, machine learning methods have received much attention and have shown great potential for improving cancer diagnosis. Most of the research works apply the machine learning methods only to image data or sometimes only to tabular data\cite{xia2023predicting}. Obviously, the limitation of the information source can affect the diagnosis performance. In this paper, we present a baseline for HCC diagnosis with both image data and tabular data.


\section{Materials and Methods}

\subsection{QHCC dataset}
The QHCC dataset consists of clinical data collected from 100 patients with HCC. The clinical data include CT, MRI, clinical laboratory test reports (see appendix), and research electronic data capture (REDCap)\cite{patridge2018research} case report forms collected before and after hepatectomy.
The lesions in both the venous and arterial phases of 179 contrast-enhanced T1w MRI and 127 contrast-enhanced CT images are segmented by a medical professional.


In this paper, we consider TNM staging\cite{subramaniam2013review} as the classification task. 95 patients with TNM staging labels are selected. For the training and validation of the machine learning model, five rounds of patient-level random permutation cross-validation are performed, where the training and test sets are assigned at a ratio of $4:1$. 

The tabular data consists of multiple columns with each column corresponding to a variable. The procedure of filtering or selecting columns of Redcap data is detailed in Appendix A.
The information of the selected variables for the lab data is shown in Appendix B.

\subsection{Data preprocessing}

For the tabular data, the non-numerical elements are converted to numerical labels.
The missing values are replaced using a mean value for each variable.

\begin{table*}
\caption{The patient distribution of the TNM stage labels of the QHCC dataset.}
\label{table}
\setlength{\tabcolsep}{3pt}
\centering
\centerline{\resizebox{8.0cm}{!}{ 
		\begin{threeparttable}[b]
\begin{tabular}{|c|c|c|c|c|c|c|c|c|}
\hline
Labels             & T0 & T1 & T2 & T3 & T4 & TX & Null & Total \\ \hline
Number of Patients & 1  & 53 & 22 & 13 & 5  & 1  & 5    & 100   \\ \hline
\end{tabular}
\end{threeparttable}}}
\label{tab-dist}
\end{table*}

As for the target prediction label, we observe that the distribution of the T-stage labels is highly unbalanced as shown in Table \ref{tab-dist}. To better balance the label distribution, we merge the labels "T0", "T1", and "TX" into the label "TX, T0 or T1", and "T3" and "T4" into the label "T3 or T4".

As each patient has multiple contrast-enhanced MRI and CT images, we augment the data using different combinations of MRI and CTs with the Redcap and Lab data. In this way, we get 376 samples, each consisting of an MRI image, a CT image, and the corresponding Redcap and lab data.

\subsection{Radiomics feature extraction}
The largest lesions in contrast-enhanced MRI and CT images are considered. We use PyRadiomics\cite{van2017computational} to extract the radiomics features from the segmented lesions. The radiomics feature classes considered include shape-based first-order statistics (3D), gray level co-occurrence matrix (GLCM), gray level run length matrix (GLRLM), gray level size zone matrix (GLSZM), gray level dependence matrix (GLDM), and neighboring gray tone difference matrix (NGTDM).

\subsection{Feature selection}
We select related features from the preprocessed tabular data and the extracted radiomics features based on the training set. The mutual information\cite{shannon1948mathematical} for the discrete target variable is estimated using Sklearn functions\cite{scikit-learn}. At least half of the features are selected to build the input vectors for the machine learning classifier. We adopt the methods of recursive feature elimination with cross-validation\cite{guyon2002gene} to determine the number of features to be selected for the given training samples. 


\subsection{Classifier training and test}
We adopt XGBoost\cite{chen2016xgboost} as the classifier, which is one of the most efficient gradient-boosted decision trees (GBDTs) and also the state-of-the-art machine learning method for tabular data\cite{mcelfresh2024neural}. We use the softmax objective. The evaluation loss is the cross-entropy loss for multi-class.

\section{Results}
We consider two metrics: the prediction accuracy (ACC) and the area under the curve (AUC). The "mean $\pm$ std" is reported. Table \ref{tab-fivefolds} shows the ACC and AUC results for inputs of different combinations. It shows that Redcap data allows for higher prediction accuracy than other data. Adding lab data can improve the performance further. Both the radiomics features extracted from CT and MRI can help to improve the prediction based on Redcap data or lab data. CT radiomics features work better with Redcap data for prediction.
The higher classification accuracy is achieved by combining MR radiomics features with Redcap and lab data. When using all the tabular data and radiomics features, we get the smallest prediction std with the second-best AUC score.

\begin{table*}
\caption{The classification results on the QHCC dataset.}
\label{table}
\setlength{\tabcolsep}{3pt}
\centering
\centerline{\resizebox{18.0cm}{!}{ 
		\begin{threeparttable}[b]
\begin{tabular}{|c|cc|cc|cc|cc|}
\hline
         & \multicolumn{2}{c|}{Redcap data}               & \multicolumn{2}{c|}{Lab data}                  & \multicolumn{2}{c|}{Redcap $+$ Lab data}         & \multicolumn{2}{c|}{Null}                      \\ \hline
         & \multicolumn{1}{c|}{ACC}         & AUC         & \multicolumn{1}{c|}{ACC}         & AUC         & \multicolumn{1}{c|}{ACC}         & AUC         & \multicolumn{1}{c|}{ACC}         & AUC         \\ \hline
CT       & \multicolumn{1}{c|}{0.88  $\pm$ 0.04} & 0.90  $\pm$ 0.06 & \multicolumn{1}{c|}{0.51  $\pm$ 0.08} & 0.68  $\pm$ 0.08 & \multicolumn{1}{c|}{0.83  $\pm$ 0.07} & 0.90  $\pm$ 0.02 & \multicolumn{1}{c|}{0.55  $\pm$ 0.08} & 0.64  $\pm$ 0.06 \\ \hline
MRI      & \multicolumn{1}{c|}{0.83  $\pm$ 0.12} & 0.88  $\pm$ 0.09 & \multicolumn{1}{c|}{0.49  $\pm$ 0.06} & 0.65  $\pm$ 0.10 & \multicolumn{1}{c|}{0.89  $\pm$ 0.05} & 0.93  $\pm$ 0.03 & \multicolumn{1}{c|}{0.51  $\pm$ 0.10} & 0.57  $\pm$ 0.08 \\ \hline
CT $+$ MRI & \multicolumn{1}{c|}{0.88  $\pm$ 0.05} & 0.90  $\pm$ 0.07 & \multicolumn{1}{c|}{0.55  $\pm$ 0.04} & 0.69  $\pm$ 0.09 & \multicolumn{1}{c|}{0.86  $\pm$ 0.03} & 0.92  $\pm$ 0.02 & \multicolumn{1}{c|}{0.54  $\pm$ 0.09} & 0.60  $\pm$ 0.03 \\ \hline
Null     & \multicolumn{1}{c|}{0.82  $\pm$ 0.12} & 0.88  $\pm$ 0.12 & \multicolumn{1}{c|}{0.48  $\pm$ 0.06} & 0.50  $\pm$ 0.06 & \multicolumn{1}{c|}{0.85  $\pm$ 0.07} & 0.90  $\pm$ 0.04 & \multicolumn{1}{c|}{Null}        & Null        \\ \hline
\end{tabular}
\end{threeparttable}}}
\label{tab-fivefolds}
\end{table*}





\begin{figure*}[!htbp]	\centerline{\includegraphics[width=18.0cm]{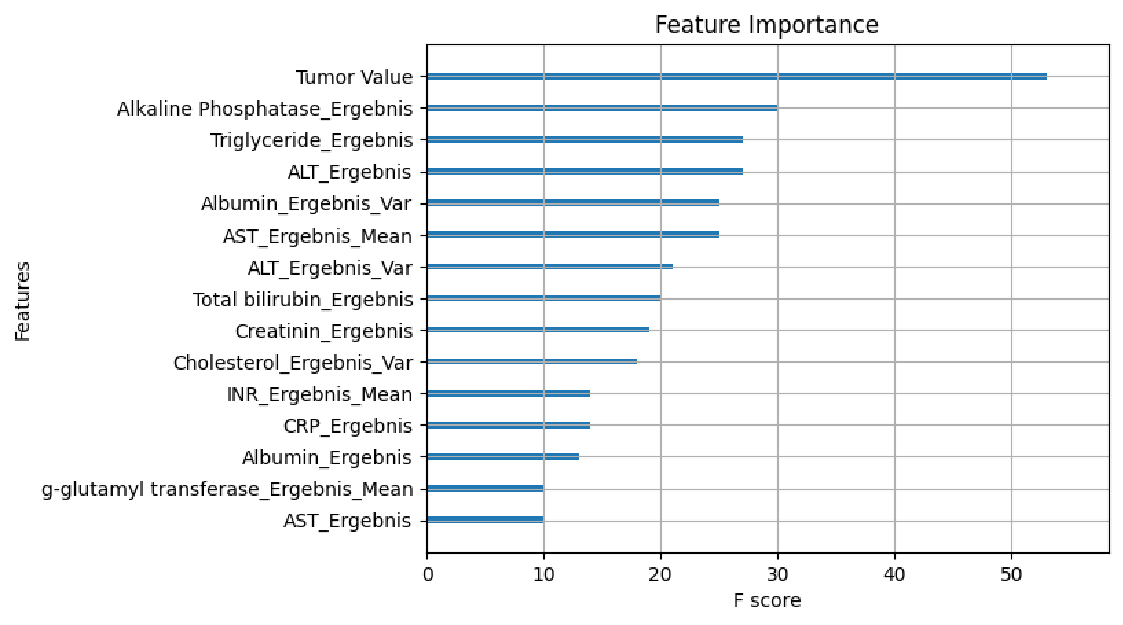}}
	\caption{The importance of selected tabular features for TNM staging.}
	\label{fig_feature_importance}
\end{figure*}

\begin{figure*}[!htbp]	\centerline{\includegraphics[width=18.0cm]{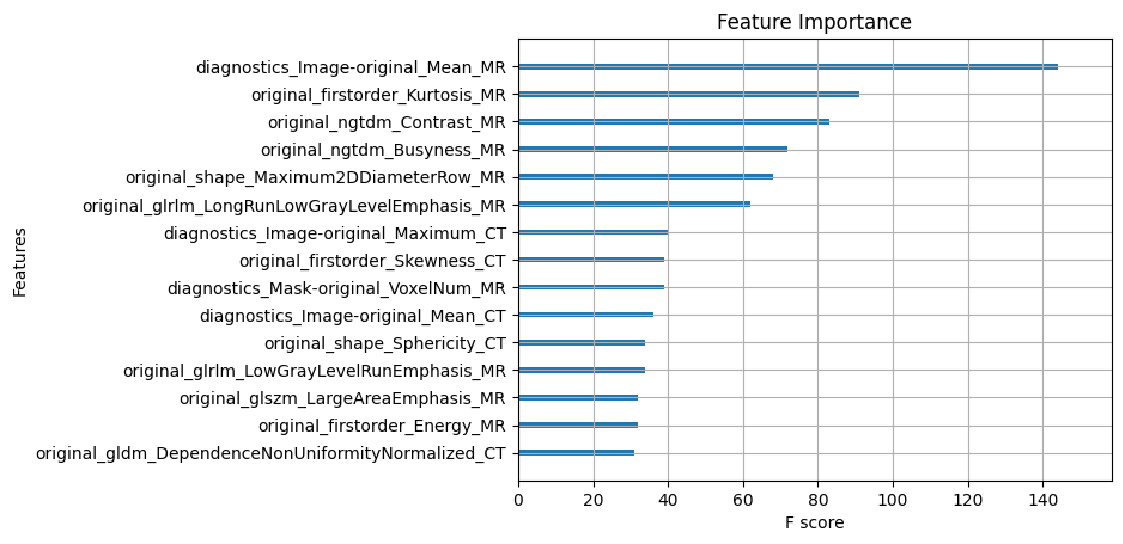}}
	\caption{The importance of selected radiomics features for TNM staging.}
	\label{fig_feature_importance_rad}
\end{figure*}

Fig. \ref{fig_feature_importance} shows the importance of selected tabular features considering the experiment with a random seed using only Redcap and lab data. Fig. \ref{fig_feature_importance_rad} shows the importance of radiomics features when doing prediction with only radiomics features.

\begin{figure*}[!htbp]	\centerline{\includegraphics[width=18.0cm]{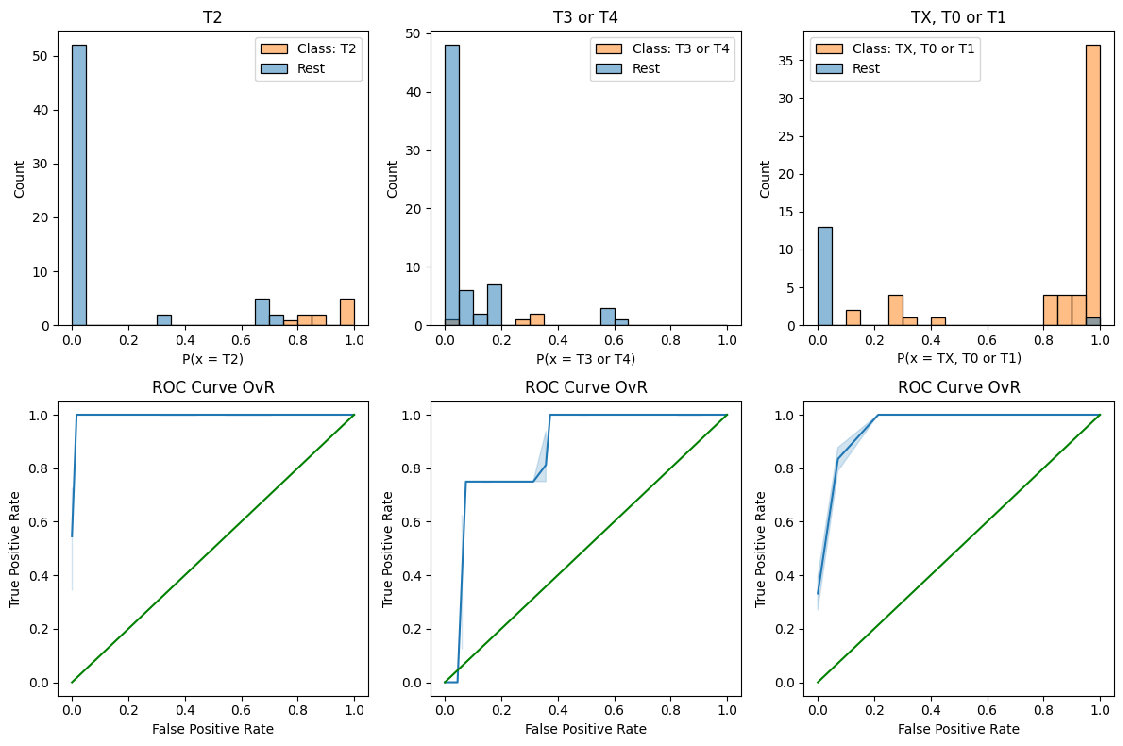}}
	\caption{The probability distributions and the ROC curves (One vs Rest, OvR). In the first row are results of the probability distributions. The second row shows the corresponding ROC curves.}
	\label{fig_rocs}
\end{figure*}

Fig. \ref{fig_rocs} shows the probability distributions of the predictions as well as the corresponding one-vs-rest receiver operating characteristic (ROC) curves for experiments with all the data modalities.

From the results, we can see that the tabular data-only or the radiomics-feature-only prediction may not give a favorable prediction accuracy. The radiomics features related to the tumor size and shape, the age of the patients,  and "Tumor value" variable in the tabular data are frequently selected as highly related features for TNM staging, which is reasonable and is demonstrated from the feature importance shown in Fig. \ref{fig_feature_importance}. 
The combination of both tabular data and radiomics features gives the highest prediction performance. 

\FloatBarrier

\section{Conclusion}

In this paper, we present a baseline model for handling multimodal clinical data classification using the state-of-the-art tabular data classifier. Our experiments demonstrate the effectiveness of combining tabular data and radiomics image features (from contrast-enhanced CT and MRI) for predicting TNM staging of HCC. 


\section{Acknowledgment}
This project was funded by BMBF grant Q-HCC \#grant number\#.

\appendix
\section{Process of Filtering Feature Columns (Redcap data)}

When filtering and selecting feature columns from the redcap dataset, the following principles were applied to ensure data simplicity, relevance, and model effectiveness:
\subsection{Filter out features with less than 80\% non-missing values}
Columns with a non-missing value proportion below 80\% were excluded to reduce noise and minimize the impact of missing data. This step ensures the retention of features with high information density, thereby improving data quality and model training efficiency.

\subsection{Filter out columns with identical values across all samples}
Columns with the same value across all samples were removed because they contribute nothing to the classification task.

\subsection{Filter out data not closely related to the patient’s cancer condition}
Features unrelated to liver cancer classification, such as the following, were removed: Type of data collection (e.g., retrospective or prospective); Patient nationality; Categories of imaging diagnosis (e.g., whether a specific imaging study was performed); Examination dates; Label codes (e.g., laboratory study label codes); Patient consent information; Family health information.

\subsection{Filter intermediate scores used to calculate final stage ratings}
For features associated with specific staging systems (e.g., BCLC staging with five stages), we retained only the final staging results and removed intermediate scores used for their calculation to avoid redundancy and reduce the potential for data overlap in model training.

\subsection{Filter out information related to subsequent treatment interventions}
Data related to surgical stages or treatment interventions are primarily used for assessing treatment outcomes and are not essential for liver cancer classification task. Examples include: Specific surgical stages and procedures (e.g., liver segment resection information); Types and timing of follow-up treatments (e.g., whether specific drugs were administered).

\subsection{Handle features with high similarity rates}
For columns with over 95\% identical values, features not considered critical
for classification (based on domain knowledge) or statistically insignificant to the target variable were filtered out.

\section{The selected variables for the lab data}
\begin{table*}[!htbp]
\caption{The selected variables of lab data of the QHCC dataset.}
\label{tab_labdata_var}
\setlength{\tabcolsep}{3pt}
\centering
\centerline{\resizebox{18.0cm}{!}{ 
		\begin{threeparttable}[b]
\begin{tabular}{|c|c|c|}
\hline
Variable                                & Definition                                                                                          & Range of values                \\ \hline
Hemoglobin                              & Protein in red blood cells that carries oxygen.                                                     & 13.2–16.7 g/dL                 \\ \hline
Hematocrit                              & Proportion of red blood cells in the blood.                                                         & 33.4-46.2\%                    \\ \hline
Creatinine                              & Waste product from muscle metabolism, used to assess kidney function.                               & 0.7–1.3 mg/dL                  \\ \hline
Sodium (Natrium)                        & Electrolyte important for fluid balance and nerve function.                                         & 136–145 mmol/L                 \\ \hline
ALT (Alaninaminotransferase)            & Liver enzyme used to assess liver health.                                                           & 7–40 U/L                       \\ \hline
Gamma-glutamyl transferase (GGT)        & Enzyme involved in liver and bile duct function.                                                    & 0–73 U/L                       \\ \hline
C-Reactive Protein (CRP)                & Marker for inflammation in the body.                                                                & 0-5 mg/dL                      \\ \hline
Total Bilirubin                         & Breakdown product of hemoglobin, assesses liver function.                                           & 0.2–1.0 mg/dL                  \\ \hline
Alkaline Phosphatase (ALP)              & Enzyme related to bile ducts; also linked to bone health.                                           & 40–130 U/L                     \\ \hline
Aspartate Aminotransferase (AST)        & Enzyme found in liver and muscles, used to assess liver damage.                                     & 13–40 U/L                      \\ \hline
INR (International Normalized Ratio)    & Standardized number for blood clotting time, based on prothrombin time.                             & 0.9–1.15                       \\ \hline
Albumin                                 & Main protein in blood plasma, maintains oncotic pressure and transports substances.                 & 35–48 g/L                      \\ \hline
Cholesterol                             & Type of fat found in the blood, important for cell membranes and hormones.                          & \textless{}200 mg/dL           \\ \hline
Triglycerides                           & Type of fat (lipid) found in the blood.                                                             & \textless{}150 mg/dL           \\ \hline
\multirow{3}{*}{Hemoglobin A1c (HbA1c)} & \multirow{3}{*}{Average blood glucose level over the past 2–3 months, used in diabetes monitoring.} & 4–5.6\% (normal)               \\ \cline{3-3} 
                                        &                                                                                                     & 5.7–6.4\% (pre-diabetes)       \\ \cline{3-3} 
                                        &                                                                                                     & \textgreater{}6.5\% (diabetes) \\ \hline
Alpha-fetoprotein (AFP)                 & Protein produced by the liver; elevated in liver diseases and certain cancers.                      & 0-8.1 ng/mL                    \\ \hline
Direct Bilirubin                        & Conjugated bilirubin, measures liver’s ability to excrete waste.                                    & 0.0–0.3 mg/dL                  \\ \hline
Ferritin                                & Protein that stores iron, used to assess iron levels.                                               & 26–388 ng/mL                   \\ \hline
\end{tabular}
\end{threeparttable}}}
\end{table*}    
\bibliographystyle{unsrtnat}
\bibliography{ref_qhcc}

\begin{thebibliography}{12}
\providecommand{\natexlab}[1]{#1}
\providecommand{\url}[1]{\texttt{#1}}
\expandafter\ifx\csname urlstyle\endcsname\relax
  \providecommand{\doi}[1]{doi: #1}\else
  \providecommand{\doi}{doi: \begingroup \urlstyle{rm}\Url}\fi

\bibitem[Balogh et~al.(2016)Balogh, Victor~III, Asham, Burroughs, Boktour,
  Saharia, Li, Ghobrial, and Monsour~Jr]{balogh2016hepatocellular}
Julius Balogh, David Victor~III, Emad~H Asham, Sherilyn~Gordon Burroughs, Maha
  Boktour, Ashish Saharia, Xian Li, R~Mark Ghobrial, and Howard~P Monsour~Jr.
\newblock Hepatocellular carcinoma: a review.
\newblock \emph{Journal of hepatocellular carcinoma}, pages 41--53, 2016.

\bibitem[Hennedige and Venkatesh(2013)]{hennedige2013imaging}
Tiffany Hennedige and Sudhakar~Kundapur Venkatesh.
\newblock Imaging of hepatocellular carcinoma: diagnosis, staging and treatment
  monitoring.
\newblock \emph{Cancer Imaging}, 12\penalty0 (3):\penalty0 530, 2013.

\bibitem[Abhyankar et~al.(2012)Abhyankar, Demner-Fushman, and
  McDonald]{abhyankar2012standardizing}
Swapna Abhyankar, Dina Demner-Fushman, and Clement~J McDonald.
\newblock Standardizing clinical laboratory data for secondary use.
\newblock \emph{Journal of biomedical informatics}, 45\penalty0 (4):\penalty0
  642--650, 2012.

\bibitem[Patridge and Bardyn(2018)]{patridge2018research}
Emily~F Patridge and Tania~P Bardyn.
\newblock Research electronic data capture (redcap).
\newblock \emph{Journal of the Medical Library Association: JMLA}, 106\penalty0
  (1):\penalty0 142, 2018.

\bibitem[Xia et~al.(2023)Xia, Zhou, Meng, Zha, Yu, Wang, Song, Wang, Tang, Xu,
  et~al.]{xia2023predicting}
Tian-yi Xia, Zheng-hao Zhou, Xiang-pan Meng, Jun-hao Zha, Qian Yu, Wei-lang
  Wang, Yang Song, Yuan-cheng Wang, Tian-yu Tang, Jun Xu, et~al.
\newblock Predicting microvascular invasion in hepatocellular carcinoma using
  ct-based radiomics model.
\newblock \emph{Radiology}, 307\penalty0 (4):\penalty0 e222729, 2023.

\bibitem[Subramaniam et~al.(2013)Subramaniam, Kelley, and
  Venook]{subramaniam2013review}
Somasundaram Subramaniam, Robin~K Kelley, and Alan~P Venook.
\newblock A review of hepatocellular carcinoma (hcc) staging systems.
\newblock \emph{Chinese clinical oncology}, 2\penalty0 (4):\penalty0 33--33,
  2013.

\bibitem[Van~Griethuysen et~al.(2017)Van~Griethuysen, Fedorov, Parmar, Hosny,
  Aucoin, Narayan, Beets-Tan, Fillion-Robin, Pieper, and
  Aerts]{van2017computational}
Joost~JM Van~Griethuysen, Andriy Fedorov, Chintan Parmar, Ahmed Hosny, Nicole
  Aucoin, Vivek Narayan, Regina~GH Beets-Tan, Jean-Christophe Fillion-Robin,
  Steve Pieper, and Hugo~JWL Aerts.
\newblock Computational radiomics system to decode the radiographic phenotype.
\newblock \emph{Cancer research}, 77\penalty0 (21):\penalty0 e104--e107, 2017.

\bibitem[Shannon(1948)]{shannon1948mathematical}
Claude~Elwood Shannon.
\newblock A mathematical theory of communication.
\newblock \emph{The Bell system technical journal}, 27\penalty0 (3):\penalty0
  379--423, 1948.

\bibitem[Pedregosa et~al.(2011)Pedregosa, Varoquaux, Gramfort, Michel, Thirion,
  Grisel, Blondel, Prettenhofer, Weiss, Dubourg, Vanderplas, Passos,
  Cournapeau, Brucher, Perrot, and Duchesnay]{scikit-learn}
F.~Pedregosa, G.~Varoquaux, A.~Gramfort, V.~Michel, B.~Thirion, O.~Grisel,
  M.~Blondel, P.~Prettenhofer, R.~Weiss, V.~Dubourg, J.~Vanderplas, A.~Passos,
  D.~Cournapeau, M.~Brucher, M.~Perrot, and E.~Duchesnay.
\newblock Scikit-learn: Machine learning in {P}ython.
\newblock \emph{Journal of Machine Learning Research}, 12:\penalty0 2825--2830,
  2011.

\bibitem[Guyon et~al.(2002)Guyon, Weston, Barnhill, and Vapnik]{guyon2002gene}
Isabelle Guyon, Jason Weston, Stephen Barnhill, and Vladimir Vapnik.
\newblock Gene selection for cancer classification using support vector
  machines.
\newblock \emph{Machine learning}, 46:\penalty0 389--422, 2002.

\bibitem[Chen and Guestrin(2016)]{chen2016xgboost}
Tianqi Chen and Carlos Guestrin.
\newblock Xgboost: A scalable tree boosting system.
\newblock In \emph{Proceedings of the 22nd acm sigkdd international conference
  on knowledge discovery and data mining}, pages 785--794, 2016.

\bibitem[McElfresh et~al.(2024)McElfresh, Khandagale, Valverde, Prasad~C,
  Ramakrishnan, Goldblum, and White]{mcelfresh2024neural}
Duncan McElfresh, Sujay Khandagale, Jonathan Valverde, Vishak Prasad~C, Ganesh
  Ramakrishnan, Micah Goldblum, and Colin White.
\newblock When do neural nets outperform boosted trees on tabular data?
\newblock \emph{Advances in Neural Information Processing Systems}, 36, 2024.

\end{thebibliography}

\end{document}